\documentclass[runningheads]{llncs}
\usepackage{graphicx}
\usepackage[T1]{fontenc}
\usepackage{amsmath}
\usepackage{amssymb}
\usepackage{booktabs}
\usepackage[mathscr]{euscript}
\usepackage{xcolor}
\usepackage{multirow}
\newcommand{\Tau}{\mathrm{T}}

\begin{document}
\title{Target-driven One-Shot Unsupervised Domain Adaptation}
\author{Julio Ivan Davila Carrazco\inst{1,2}
\and Suvarna Kishorkumar Kadam\inst{1} 
\and Pietro Morerio\inst{1} 
\and Alessio Del Bue\inst{1} 
\and Vittorio Murino\inst{1,3}}
\institute{Pattern Analysis and Computer Vision (PAVIS), Italian Institute of Technology, Genoa, Italy 
\and Department of Marine, Electrical, Electronic and Telecommunications Engineering, University of Genoa, Genoa, Italy 
\and Department of Computer Science, University of Verona, Verona, Italy \\
\email{\{julio.davila, suvarna.kadam, pietro.morerio, alessio.delbue, vittorio.murino\}@iit.it}}

\authorrunning{J.I.D. Carrazco et al.}
\date{May 2023}
\maketitle              % typeset the header of the contribution
\begin{abstract}
%The abstract should briefly summarize the contents of the paper in
%%% NEW ABSTRACT
In this paper, we introduce a novel framework for the challenging problem of One-Shot Unsupervised Domain Adaptation (OS-UDA), which aims to adapt to a target domain with only a single unlabeled target sample. Unlike existing approaches that rely on large labeled source and unlabeled target data, our Target-driven One-Shot UDA (TOS-UDA) approach employs a learnable augmentation strategy guided by the target sample's style to align the source distribution with the target distribution. Our method consists of three modules: an augmentation module, a style alignment module, and a classifier. Unlike existing methods, our augmentation module allows for strong transformations of the source samples, and the style of the single target sample available is exploited to guide the augmentation by ensuring perceptual similarity. Furthermore, our approach integrates augmentation with style alignment, eliminating the need for separate pre-training on additional datasets. Our method outperforms or performs comparably to existing OS-UDA methods on the Digits and DomainNet benchmarks.
%%% OLD ABSTRACT
%In this paper, we introduce a novel framework for the challenging problem of One-Shot Unsupervised Domain Adaptation (OS-UDA). Unlike existing UDA approaches, which leverage sufficiently large labeled source and unlabeled target data, our approach attempts to adapt to the target domain when only a single unlabeled target sample is available. Such adaptation setting is clearly quite challenging. To address this task, we propose a novel UDA approach, named Target-driven One-Shot UDA (TOS-UDA), which employs a learnable augmentation strategy guided by the \textit{style} of the single target sample to gradually align the source distribution with the target one. Our method is composed of three modules: an augmentation module, a style alignment module, and a classifier. Compared with existing adversarial augmentation methods, our augmentation module is guided by the target's style and allows for strong transformations of the source samples. Unlike existing OS-UDA methods, our adaptation method integrates augmentation with style alignment and  does not need  separate pre-training of style transfer module on additional dataset. Our method is tested on  Digits and DomainNet benchmarks and perform better or similar compared with the existing OSUDA methods.
\end{abstract}

\section{Introduction}
\label{sec:intro}
%------------------------------------------------------------------------
\begin{figure}[h]
\centering
\includegraphics[width=0.6\textwidth]{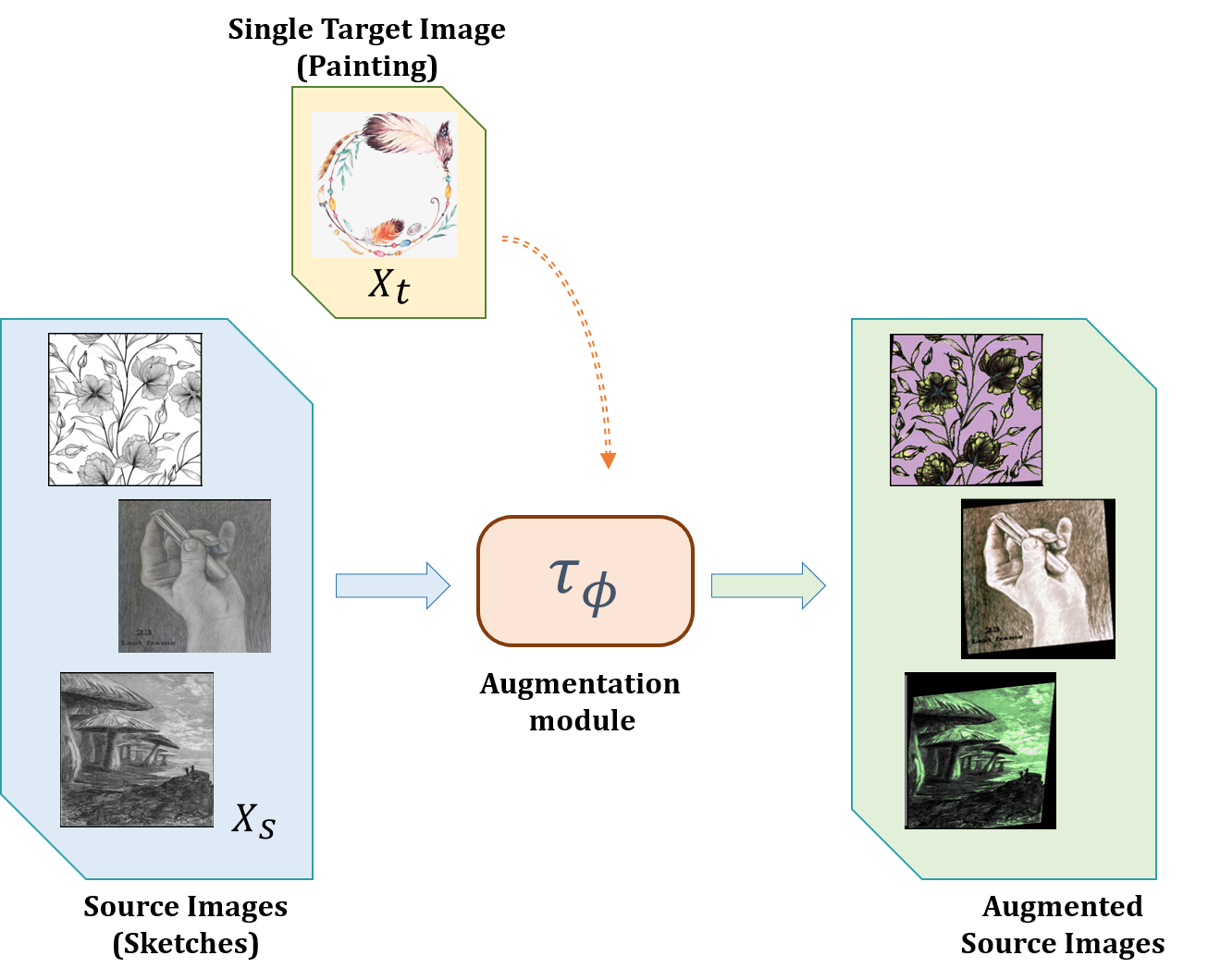}
\caption{Target-driven augmentation of source samples with TOS-UDA. $\tau$ is a learnable augmentation function with parameters $\phi$ that transforms the source images into target-like images (Sketches to Painting in this case). 
}
\label{fig:AugmentedImgs}
\end{figure}
Training deep learning models typically requires the availability of a large amount of data, which also represents a core problem for applications in which such information is not easily accessible. 
%In actual scenarios, data (e.g., images, videos, audio signals, etc.) is in fact difficult to obtain and, in some cases, even impossible to acquire in large quantities. 
Hence, there is a need for methods to effectively learn from either fewer or alternative data that is easily accessible.
%, yet has similarities to the original data.
The situation becomes even more challenging when we have to deal with different ``domains'', i.e., when data distributions in training (source domain) and test (target domain) are different. How to exploit such different data for a better generalization in the target (test) domain is an active topic in computer vision nowadays, with plenty of methods proposed in the literature. 
The term `domain' in this context is not clearly defined, and could roughly be identified with ``a set of features'', such as color or texture, characterizing samples of a specific dataset. The typical task here is to properly learn from the \textit{labeled} source data and the \textit{unlabeled} target data, i.e., how to transfer knowledge from source to target, the latter being also our test scenario. This is called unsupervised domain adaptation (UDA).
 %%% CUT TO REACH LIMIT %%%
%Depending on the source and target domain characteristics this transfer can be more or less difficult, but the underlying assumption is that we do have enough data (in both source and target datasets) to train our model.
 %%% CUT TO REACH LIMIT %%%
A much less investigated, but more challenging task is to explore what happens when the source and/or target data are just a few samples, or even just a single data item.
In these cases, we do not have enough statistics to capture the data variability, and then to properly generalize to new data. Hence, we might resort to methods to generate new or surrogate data from the few samples available in order to execute proper training.
These UDA scenarios are named Few-Shot UDA (FS-UDA) \cite{yue2021prototypical,9859804} or, to an extreme case, One-Shot UDA (OS-UDA) \cite{luo2020adversarial,gu2022few}. FS-UDA denotes the presence of a few (labeled) samples in the source domain only, while there is enough unlabeled data in the target domain. OS-UDA instead assumes the presence of only one (unlabeled) sample in the target domain. FS-UDA and OS-UDA are difficult to learn tasks due to insufficient training data. 
 %%% CUT TO REACH LIMIT %%%
%Empirical risk minimization is challenging to implement for Few-shot settings ~\cite{wang2020generalizing}, and some form of prior knowledge needs to be roped in to improve the generalization despite data scarcity. 
%For OS-UDA or FS-UDA methods, prior knowledge about the shared label space between source and target domain is available and these methods should exploit this knowledge along with additional domain characteristics to learn adaptation. 
 %%% CUT TO REACH LIMIT %%%
Existing methods for solving FS-UDA, OS-UDA are quite limited. To the best of our knowledge, there is just one OS-UDA method, ASM \cite{luo2020adversarial} that performs UDA for classification and segmentation tasks. 

In this paper, we introduce a new OS-UDA method where a {single} target sample is available to align the source domain with the target domain. We design a novel augmentation method implementing gradual transformations of the source images driven by the similarity with target's style. Our augmentation strategy achieves diverse, but controlled augmentations using a custom adversarial loss that rewards stronger augmentations but, at the same time, also penalizes too large deviations from the target's style. These diverse augmentations finally support robust adaptation to the target domain. 
In summary, our main contributions are as follows.
\begin{itemize}
    \item We present a novel approach to tackle OS-UDA task, addressing the extreme case of target domain adaptation when only one unlabeled target sample is available.  Our method performs target-driven augmentation of source images in an adversarial manner, and is the first end-to-end OS-UDA method that does not require specially pretrained style transfer modules.
    \item We guide source image augmentations with a style alignment module that controls the augmentation parameters while  enforcing the style similarity between source and target domains. 
    \item We report comparable, and in few cases, state-of-the-art results for OS-UDA on Digits and DomainNet benchmarks. 
\end{itemize}
 %%% CUT TO REACH LIMIT %%%
%The rest of this paper is organized as follows. Section 2 presents a review of the literature regarding UDA, FS-UDA, OS-UDA and augmentation methods, outlining their main characteristics and known limitations. Section 3 presents our proposed framework focusing on each individual component and illustrating the final training strategy. Section 4 presents the benchmarks and baselines used for evaluating our approach, as well as  the results of our experiments together with an ablation study. Finally, Section 5 summarizes our proposed approach and reports some conclusive remarks.
 %%% CUT TO REACH LIMIT %%%
The rest of this paper is structured as follows: Section 2 reviews the literature on UDA, FS-UDA, OS-UDA, and augmentation methods. Section 3 presents our proposed framework and training strategy. Section 4 outlines the benchmarks, baselines, experimental results, and ablation study. Finally, Section 5 summarizes our approach and provides concluding remarks.
\section{Related Work}
\label{sec:related_work}
Unsupervised domain adaptation (UDA) methods leverage the available data for learning and can face different scenarios where either the source, target or both domains have scarce data. Data augmentation is often the \textit{de facto} approach to address data scarcity, as it employs diverse transformation strategies to increase the variability of available data. Our proposed approach is based on a novel adversarial augmentation strategy for UDA, and therefore we briefly discuss related work in both areas. 
\subsection{Data Augmentation}
Image augmentation methods are broadly categorized as model-free, model-based, and optimization-based \cite{xu2022comprehensive}. The model-free methods use classic image processing to alter geometric or color 
information. Geometric transformations \cite{taylor2018improving} are often applied to a single image by flipping, rotating, or cropping it. 
 %%% CUT TO REACH LIMIT %%%
 %Patch-based augmentation methods~\cite{terrance2017cutout,zhong2020random,gridmask2020} are successful strategies that augment by masking parts of an image. Geometric and color transformations are commonly applied to single images, but can also be applied to multiple images.  Specifically, such augmentation methods blend input images to generate new samples (e.g., \cite{zhang2017mixup}). Several mixing strategies are proposed, e.g.,  \textit{mixup}~\cite{zhang2017mixup}, which applies linear interpolation in both data and the corresponding labels, or \cite{inoue2018data}, which combines the samples with a common label. CutMix~\cite{yun2019cutmix} and Mosaic augmentation in yolov4~\cite{bochkovskiy2020yolov4} strategically fuse the patches of the image for training and achieve better generalization.
 %%% CUT TO REACH LIMIT %%%
 %%% CUT TO REACH LIMIT %%%
 %GridMix~\cite{baek2021gridmix} combines patch-based and mixup methods to fuse patches of a single image. 
 %Other mixup methods, such as PuzzleMix~\cite{kim2020puzzle} and SuperMix~\cite{Dabouei2021SuperMixST}, process images to find important regions such as image foreground faster, and then apply mixup on these regions.
%%% CUT TO REACH LIMIT %%%
Model-based image augmentation methods pre-train generative models such as GANs~\cite{mirza2014generative,goodfellow2020generative} and variants, to generate images. The pre-trained model may generate images unconditionally or subject to certain conditions. The distribution of generated images is expected to be similar to the original dataset. 
%%% CUT TO REACH LIMIT %%%
%Conditional GANs~\cite{mirza2014conditional} force the generator and discriminator to learn a representation that can generate images conditioned on class labels. 
%AugGAN~\cite{huang2018auggan} is a structure-aware image-to-image translation network that can augment while preserving label consistency. 
%%% CUT TO REACH LIMIT %%%
 %%% CUT TO REACH LIMIT %%%
%When an image is used as a conditioning factor instead of a label, such augmentation methods are called image-to-image translations. As an image consists of both content and style, it is possible to separate content that is label-dependent, and style which is independent of a class label.
 %%% CUT TO REACH LIMIT %%%
The optimization-based methods implement trainable augmentation strategies \cite{cubuk2019autoaugment,cubuk2020randaugment,Suzuki_2022_CVPR} that learn the transformations, and are usually preferred as they do not need to be configured for specific datasets.
While data augmentation mostly improves model generalization, it can result in transformations that are too extreme~\cite{Suzuki_2022_CVPR}. Our proposed approach is optimization-based that uses the single target image's style to control the source image augmentations. Our novel target-driven augmentation strategy is inspired by the fact that we can separate the style from the content of the available single target image, and use it to adjust the transformations to be applied to source domain images.
\subsection{Unsupervised Domain Adaptation}Most common UDA settings transfer knowledge from a label-rich source domain to unlabeled target domain, where the source and target domains share the same label space (i.e., same classes). Many UDA methods try to find a shared feature space where the overlap (confusion) between the source and target distribution is maximum. In this way, for any given sample projected in such space, it is often difficult to discriminate if it belongs to the source or target domain. 
%Model generalization is crucial to properly learn such space, and data augmentation methods are known to help improving model generalization~\cite{shorten2019survey}.\\
UDA learning strategies can be broadly categorized as self-supervised, augmentation-based or adversarial methods. UDA methods e.g. \cite{volpi2018adversarial,xu2020adversarial,chen2020adversarial} assume that only unlabeled target samples are available. 
Some UDA methods focus on expanding the available target data to train their models, e.g., Volpi et al. \cite{volpi2018adversarial} trained a domain-invariant feature extractor by applying feature augmentation using GANs. Minghao et al. \cite{xu2020adversarial} introduced Domain Mixup (DM-ADA), which uses a GAN approach and mixup \cite{zhang2017mixup} to learn a more continuous domain-invariant latent space. Some methods apply self-supervision to generate pseudo-labels for the target data \cite{chen2020adversarial}. 
%As an example, \cite{chen2020adversarial} presented Adversarial-Learned Loss for Domain Adaptation (ALDA) that combines adversarial learning and self-supervision to align the feature distribution. 
%The term Few-shot is commonly referred to a setting where data samples are limited to a small quantity (i.e., 100, 50, 10 or even less) reflecting a real-word scenario. 
 %%% CUT TO REACH LIMIT %%%
%Recent research methods \cite{motiian2017few,xu2019d,9506716,yue2021prototypical} focus especially on UDA with this setting. Motiian et al. \cite{motiian2017few} introduced FADA (Few-shot Adversarial Domain Adaptation), which takes an adversarial approach to maximize the confusion between source and target, and at the same time, align their embeddings. Similarly, d-SNE~\cite{xu2019d} creates a joint embedding space by minimizing the intra-class distance while maximizing the inter-class distance \cite{xu2019d}. 
 %%% CUT TO REACH LIMIT %%%
Moreover, Few-Shot Unsupervised Domain Adaption (FS-UDA) operates in a UDA setting where unlabeled target images are accessible for training and the number of labeled samples in the source domain is small \cite{yue2021prototypical,9859804}. 
%The term Few-shot is commonly referred to a setting where data samples are limited to a small quantity (i.e., 100, 50, 10 or even less) reflecting a real-word scenario. 
One-shot UDA is a more constrained UDA scenario that assumes the presence of \textit{only one} (unlabeled) sample in the target domain. The literature focusing on solving OS-UDA is quite limited. Luo et al. \cite{luo2020adversarial} presented Adversarial Style Mining (ASM) that applies style transfer to generate new samples of the target domain. ASM applies domain randomization to generalize better for OS-UDA, and needs an external dataset (WikiArts) to pre-train its style transfer model.  Our approach is distinct as we do not need any external dataset to pre-train style transfer module. In our work, we exploit the perceptual similarity in style that exists between the source and target, to learn useful transformations while augmenting source samples. Unlike ASM~\cite{luo2020adversarial} the augmentations generated by our method are affected solely by target image's style. 
 %%% CUT TO REACH LIMIT %%%
%ASM's style transfer module is pretrained with additional dataset such as wikiArts and therefore such dataset is likely to have an influence on augmentations. Such influence may not be desirable for domain adaptation if the target domain's style drastically different from WikiArts or Source. ASM's training pipeline is not end-to-end and the additional dataset for pre-training needs to be manually selected based on task and characteristics of source and target dataset.
 %%% CUT TO REACH LIMIT %%%
 \section{Method}
\label{sec:proposed_method}
\begin{figure}[t]
\centering
\includegraphics[width=\textwidth]{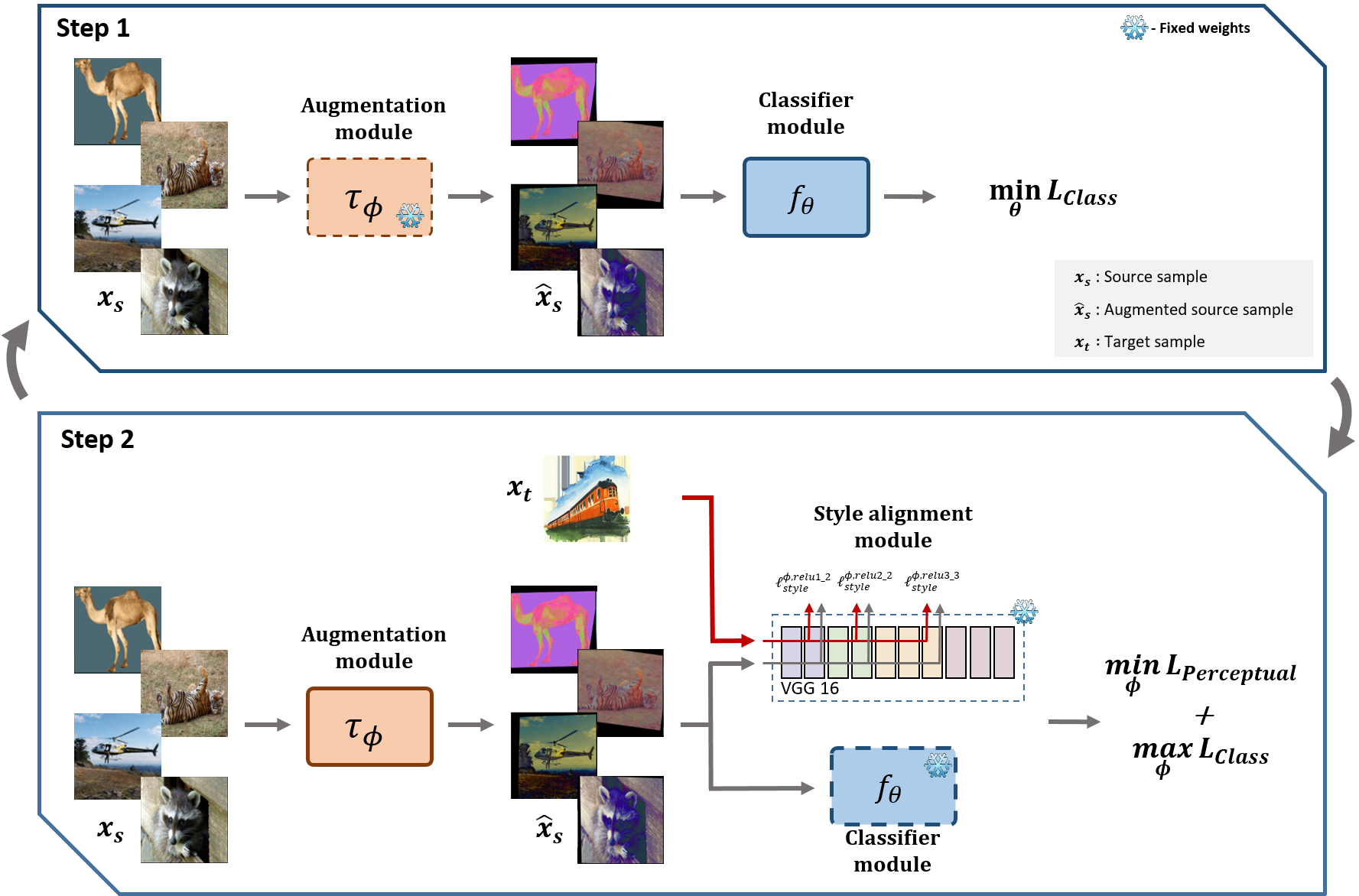}
\caption{ Training of Target-driven OS-UDA that alternatively updates the Classifier $f_\theta$ (Step-1) and Augmentation module $\Tau_\phi$ (Step-2). 
%At the start of the training, the classifier is initialized with a model trained on source-only images. 
%(Step-1) The augmentation model is frozen and the classifier is trained with classification loss on augmented samples, (Step-2) The classifier is frozen and the augmentation model is trained with a min-max loss. 
%The model learns to generate augmented samples that minimize the style loss (augmented samples  perceptually similar to the target) and maximise the classification loss  with harder samples for the classifier to classify. 
Step 1 is repeated $n$ times before step 2, where $n$ is a hyperparameter. 
}
\label{fig:FixedAugArchitecture}
\end{figure}
Adapting a model by exploiting a \textit{single} target sample is a quite challenging task. Our method tackles it by learning transformations for source domain samples to mimic the style of the target domain. A classifier trained with such augmented data is expected to generalize better to the target distribution.
Notably, the problem here is how to adapt to the target distribution when the only sample available does not constitute sufficient statistics to generate a reliable distribution. So, our envisioned solution is to implement adversarial augmentation to learn robust transformations. To this end, we introduce a novel ``style alignment'' module to guide and control the augmentations using a single target sample. 
%Unlike TeachAugment method~\cite{Suzuki_2022_CVPR}, which also implements an adversarial augmentation mechanism, our method does not need a separate teacher model for controlling the augmentations. The use of a teacher limits the range of possible augmentations. This is because the augmented samples are bounded to be correctly classified by the teacher. Therefore, the distribution of the augmented samples needs to be similar to the source distribution.
%Our style alignment module allows better control for driving the augmentations towards a specific target.
This guiding mechanism is not limited by the source distribution. Consequently, the augmented samples will be able to resemble the target distribution. In consequence, a classifier trained on augmented samples will be able to classify correctly samples from the target domain.

Our proposed model is composed of three modules: an augmentation module, a style alignment module, and a classifier. 
The augmentation module tries to learn the parameters of a set of transformations (e.g., color shift, geometrical transformation) that are applied to the source samples (see Fig. \ref{fig:FixedAugArchitecture}, Step 1). The style alignment module helps to drive the augmentation learning process by evaluating the perceptual similarities in style %and content
between the augmented source %, the source, 
and the single target sample (see Fig. \ref{fig:FixedAugArchitecture}, Step 2). 
To train this architecture, we adopt a two-step strategy that alternates between updating the classifier and the augmentation modules (see Fig. \ref{fig:FixedAugArchitecture}). In the first step, the classifier is trained using augmented samples and a classification loss. In the second step, the augmentation module is trained by using the style alignment module and the classifier in an adversarial manner. The two-step strategy was firstly used by TeachAugment~\cite{Suzuki_2022_CVPR}.  
Unlike TeachAugment method, which also implements an adversarial augmentation mechanism, our method does not need a separate teacher model for controlling the augmentations. The use of a teacher limits the range of possible augmentations. This is because the augmented samples are bounded to be correctly classified by the teacher. Therefore, the distribution of the augmented samples needs to be similar to the source distribution.
Our style alignment module allows better control for driving the augmentations towards a specific target.
The architecture's modules and the two-step strategy will be explained further in the next subsections.
\subsection{Augmentation module}
\label{Augmentation}
\begin{figure*}[ht]
\centering
\includegraphics[width=\textwidth]{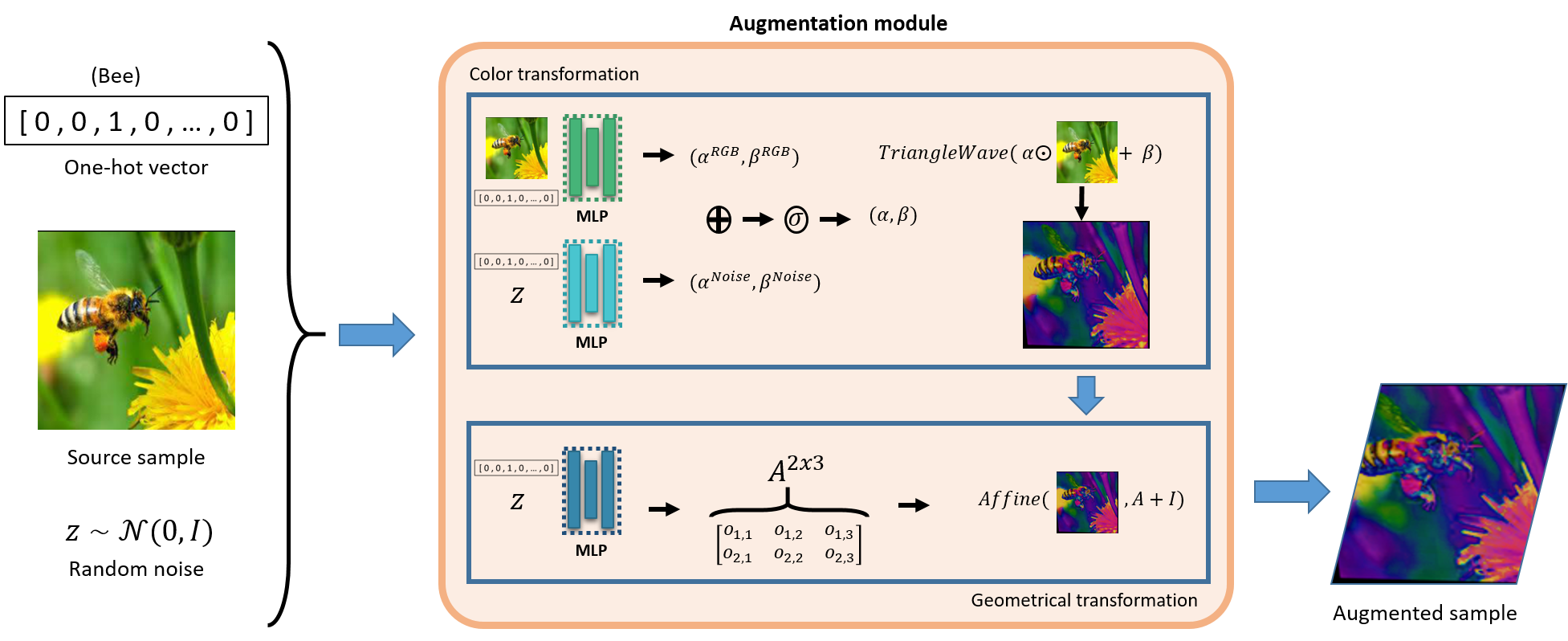}
\caption{Transformations performed within augmentation module: a) Color,  b) Geometrical transformation. Color transformations  are applied first, followed by geometrical transformation.
%Two MLPs generate the parameters of the color transformation  (scale and shift), which in turn are used as input for a triangle wave function that transforms the input image. 
%In the geometrical transformation, affine transformation matrix is generated by an MLP. This transformation is then applied to the output of the color transformation to obtain the final augmented sample.
}
\label{fig:AugmentedExpand}
\end{figure*}

The Augmentation module (AUM) is responsible for learning and applying a set of transformations to the source samples which are then used to train the classifier module. This module is composed of three Multi-Layer Perceptrons (MLPs) networks associated with two different transformations: color and geometrical transformations. AUM is trained to learn the transformations' parameters, i.e., scale and shift for the color, and an affine matrix for the geometric transformation. Hence, the augmented sample of an input image changes its appearance along the training process, and it is possible to guide the augmentation learning process to learn the parameters that will augment samples to be visually similar to a single target sample. The color transformation applied to an image $x$ is defined as:
\begin{equation}
  \hat{x} = \mbox{TriangleWave}(\alpha \odot x + \beta) 
  \hspace{0.5em} \textrm{where} \hspace{0.5em} 
\mbox{TriangleWave}(p) = \arccos(\cos(p \centerdot \pi))/\pi
  \label{eq:TriangleWave}
\end{equation}
\begin{equation}
  (\alpha, \beta) = \Tau^c_\phi(x, z, c),
\end{equation}
%\begin{equation}
%  \mbox{TriangleWave}(p) = \arccos(\cos(p \centerdot \pi))/\pi
%  \label{eq:TriangleWave}
%\end{equation}
\noindent
where $\alpha$, $\beta$ $\in$ $\mathbb{R}^3$ denote scale and shift parameters of the image color information generated by the augmentation module sub-network $\Tau^c_\phi$.
The TriangleWave function transforms the input variable $p$ into a triangular waveform in the range $[0,1]$, and $\odot$ denotes the element-wise multiplication.
The sub-network $\Tau^c_\phi$ takes as input, an image $x$, random noise $z \sim \mathscr{N}(0, I_N)$, where $\mathscr{N}(0, I_N)$ is N-dimensional unit Gaussian distribution, and $c$ that represents one-hot vector encoding for class label that serves as a context for the module (see Fig. \ref{fig:AugmentedExpand}, Color transformation). 
%This sub-network is comprised of two MLPs that generate the parameters for the transformation (see Fig. \ref{fig:AugmentedExpand}, Color transformation). 
The geometrical module applies an affine transformation to augment the input data. The transformation is defined as:
 \begin{eqnarray}
  \hat{x} = \mbox{Affine}(x, A+I) \hspace{0.5em} \textrm{, where} \hspace{0.5em} A = \Tau^g_\phi(z,c)  
    \label{eq:geometric_augmentation}
 \end{eqnarray}
%
% \begin{minipage}{0.45\textwidth}
% \begin{equation}
%   \hat{x} = \mbox{Affine}(x, A+I), 
%   \label{eq:geometric_augmentation}
% \end{equation}
% \end{minipage}
% \hfill
% \begin{minipage}{0.45\textwidth}
% \begin{equation}
%   A = \Tau^g_\phi(z,c)
%   \label{eq:geometric_augmentation}
% \end{equation}
% \end{minipage}
%%% This equations are writing above
%\begin{equation}
%  \hat{x} = \mbox{Affine}(x, A+I), 
%  \label{eq:geometric_augmentation}
%\end{equation}
%\begin{equation}
%  A = \Tau^g_\phi(z,c)
%  \label{eq:geometric_augmentation}
%\end{equation}
where Affine$(\cdot)$ denotes an affine operation applied to $x$, $A$ denotes an affine $2\times3$ matrix generated by the augmentation module $\Tau^g_\phi$, $\Tau^g_\phi(\cdot)$ is the augmentation module sub-network which is a MLP that receives as inputs, a noise vector $z \sim \mathscr{N}(0, I_N)$ and $c$, which is a one-hot encoding of the sample class, and generates the parameters of the affine transformation (see Fig. \ref{fig:AugmentedExpand}, Geometrical transformation). 
%to generate the parameters for the geometric transformation matrix $A$. 
%$\Tau^g_\phi(\cdot)$ consists of an MLP that receives as input random noise and the optional context vector to generate the parameters of the affine transformation (see Fig. \ref{fig:AugmentedExpand}, Geometrical transformation).
\subsection{Style Alignment module}
\label{StyleReg}
The purpose of the Style Alignment Module (SAM) is to drive the learning process of the augmentation module. For this module, a VGG-16 \cite{simonyan2014very} pretrained on ImageNet is used. This module leverages the style transfer methodology by exploiting a style loss. Johnson et al. \cite{johnson2016perceptual} introduced the perceptual loss to replace the traditional pixel-by-pixel loss that was used for style transfer. The perceptual loss compares a target sample, a source sample, and the augmented source sample to calculate the perceptual differences between them. This loss focuses on two aspects of the input image, its style and content.
The style refers to the texture information (e.g., color, texture, common patterns, etc.) present in the image. 
%The content focuses on preserving the semantic content, i.e. the actual objects present in the image scene.
%
%The perceptual loss is composed by the style and content loss terms, which are calculated from pairs of feature maps. 
The style loss is calculated as the squared Frobenius norm between the Gram matrices of the augmented source and the target sample. The Gram matrix serves to extract the style representation via the correlations between different filters of the extracted feature maps \cite{gatys2016image} at several network layers. The Gram matrix is defined as:
\begin{equation}
  G_{j}(x)=\dfrac{1}{C_j H_j W_j}\sum^{H_j}_{h=1}\sum^{W_j}_{w=1}\mathscr{H}_j(x) \mathscr{H}^{T}_{j}(x)
  \label{eq:gram_matrix}
\end{equation}
\noindent
where $x$ is an input image, $\mathscr{H}_j$ represents the feature map from the $j$-layer of the SAM; $C_j$, $H_j$, and $W_j$ denote the number of channels, the height, and the width of $\mathscr{H}_j$, respectively. 
The style loss is then defined as:
 \begin{eqnarray}
  L_{style} = \sum_{j}\ell^{style}_j(\Tau_\phi(x_s),x_t)
  \label{eq:style_loss} \quad \textrm{and} \quad  
    \ell^{style}_j(\hat{x_s}, x_t) = \lVert G_j(\hat{x_s}) -  G_j(x_t) \rVert^2_F
  \label{eq:style_layer_loss}
 \end{eqnarray}
% \begin{minipage}{0.40\textwidth}
% \begin{equation}
%   L_{style} = \sum_{i}\ell^{style}_i(\Tau_\phi(x_s),x_t)
%   \label{eq:style_loss}
% \end{equation}
% \end{minipage}
% \hfill
% \begin{minipage}{0.50\textwidth}
% \begin{equation}
%   \ell^{style}_j(\hat{x_s}, x_t) = \lVert G_j(\hat{x_s}) -  G_j(x_t) \rVert^2_F
%   \label{eq:style_layer_loss}
% \end{equation}
% \end{minipage}
%%% This equation is writen above
%\begin{equation}
%  \ell^{style}_j(\hat{x_s}, x_t) = \lVert G_j(\hat{x_s}) -  G_j(x_t) \rVert^2_F
%  \label{eq:style_layer_loss}
%\end{equation}
\noindent
where $x_t$ and $\hat{x_s}$ represent the target sample
and the augmented source sample, respectively. $\Tau_\phi(\cdot)$ represents the AUM with parameters $\phi$, and $j$ is the indexes of the selected layers of the SAM related to the style loss terms (more details in Section 4). 
 %%% CUT TO REACH LIMIT %%% REMOVING CONTENT LOSS %%%
%The content loss is the Euclidean distance between the features maps of the augmented source image and the original source sample. this loss is defined as:
%\begin{equation}
% \ell^{content}_j(\hat{x_s}, x_s) = \dfrac{1}{C_j H_j W_j}\lVert \mathscr{H}_j(\hat{x_s}) -  \mathscr{H}_j(x_s) \rVert^2_2
%  \label{eq:content_layer_loss}
%\end{equation}
%\noindent
%where $x_s$ and $\hat{x_s}$ represent the original source sample and the related augmented version, respectively, and $C_j$, $H_j$, $W_j$ denote the number of channels, the height, and the width of $\mathscr{H}_j$, respectively. 
 %%% CUT TO REACH LIMIT %%% REMOVING CONTENT LOSS %%%
%The features maps are extracted from specific layers of the SAM. 
The layers selection is based on the well-known fact that 
%higher layers capture better the high-level content of the image, while 
lower layers preserve the texture information better\cite{gatys2016image}.
The total style loss $L_{style}$ is the sum of the individual style %and content 
losses of the selected layers of the SAM. 
%These losses are represented as:
%%% This equation is writen above
%\begin{equation}
%  L_{style} = \sum_{i}\ell^{style}_i(\Tau_\phi(x_s),x_t) = \sum_{i}\ell^{style}_i(\hat{x_s},x_t)
%  \label{eq:style_loss}
%\end{equation}
%\begin{equation}
%  L_{content} = \sum_{j}\ell^{content}_j(\Tau_\phi(x_s),x_s) = \sum_{j}\ell^{content}_j(\hat{x_s},x_s)
%  \label{eq:content_loss}
%\end{equation}
\noindent
%where $\Tau_\phi(\cdot)$ represents the AUM with parameters $\phi$, and $i$ is the indexes of the selected layers of the SAM related to the style loss terms (more details in Section 4). 
%Finally, Eq. \ref{eq:style_loss} and Eq. \ref{eq:content_loss} are used to define the perceptual loss as follows:
%\begin{equation}
%  L_{perceptual} = \lambda_{style}L_{style} + \lambda_{content}L_{content}
%  \label{eq:perceptual_loss}
%\end{equation}
%\noindent
%Where $\lambda_{style}$ and $\lambda_{content}$ act as a weighting factors to leverage the influence of the style or content %loss in the training process.
%
\subsection{Classifier module}
\label{ClassMod}
The classifier module (CLM) has a twofold task, namely, to classify augmented samples and to collaborate with the SAM for training the AUM. The former task focuses on learning how to correctly classify the augmented source samples. The latter aims at training the augmentation module in order to learn transformations that are harder for the CLM to train on.
In this case, the classifier is paired together with SAM to perform adversarial training. For both tasks, a classification (cross-entropy, CE) loss is used in a minmax process: for the first task, the objective is to minimize the CE loss, while in the second one, it should be maximized. The classification loss is:
\begin{equation}
  L_{class} = -y_k\log f_\theta(\Tau_\phi(x_s))
  \label{eq:classification_loss}
\end{equation}
\noindent
where $x$ is a source sample, $y$ $\in$ $\{0,1\}^K$ denotes the one-hot ground-truth vector, and the number of classes is $K$, $\Tau_\phi(\cdot)$ denotes the augmentation module with parameters $\phi$, and $f_\theta(\cdot)$ represents the classifier network with parameters $\theta$.
\subsection{Two-step training process}
\label{TwoStep}
Our training method optimizes the augmentation and classifier modules end-to-end with an alternate 2-step strategy (refer to Fig. \ref{fig:FixedAugArchitecture}). In Step 1, we train the CLM, while AUM is frozen. The input for CLM is the augmented source image, $\hat{x}$, passed through the augmentation module, while SAM is not used. To optimize CLM, the classification loss (Eq.\ref{eq:classification_loss}) is minimized (supervised learning). 
%In other words, Step 1 applies supervised training for classification where the modules have access to labeled source data.
In Step 2, AUM is then optimized. Its purpose is to learn a transformation to apply onto the source samples. 
%Such transformation will allow the classifier module to generalize better to other domains in general, but privileging the target domain. 
In this step, the classifier aims at maximizing the classification loss on the augmented samples. 
By doing so, AUM will learn stronger transformations for the source samples, but this may not be sufficient to perform domain adaptation on the target. That is why our the style alignment module is introduced so as to improve the classification on the target domain. In our case, SAM drives the learning process by imposing a style loss (Eq. \ref{eq:style_loss}). 
As a result, the learned transformations will generate samples that resemble the style of the target domain. 
Finally, the augmentation module parameters $\phi$ are updated by adversarial training by maximizing the classification loss (Eq. \ref{eq:classification_loss}) and minimizing the style loss (Eq. \ref{eq:style_loss}).
\section{Experiments}
\label{sec:experiments}
\subsection{Benchmarks}
\begin{figure}[t]
\centering
\includegraphics[width=0.55\textwidth]{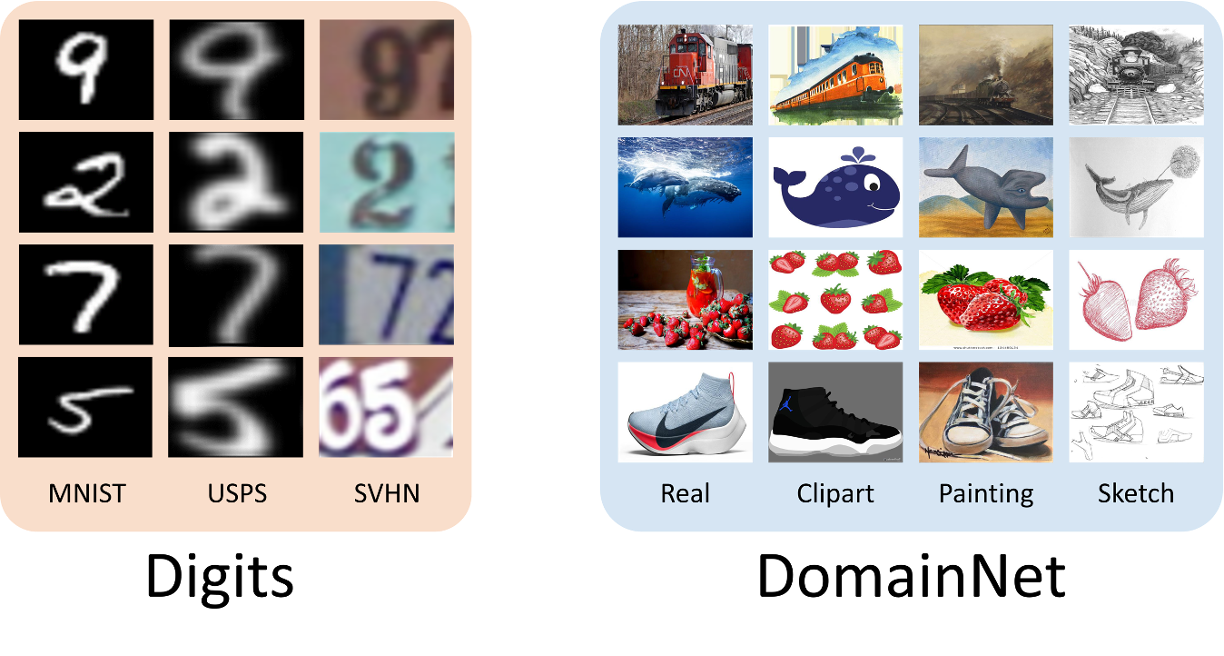}
\caption{Domain adaptation benchmarks used for evaluation of TOS-UDA: Digits with three domains (MNIST, USPS, and SVHN) and DomainNet with four domains (Real, Clipart, Painting, and Sketch).}
\label{fig:Benchmark}
\end{figure}

To evaluate our approach, we make use of two well-known DA benchmarks: Digits and DomainNet \cite{peng2019moment}. \textbf{Digits.} It is composed of three datasets: MNIST \cite{lecun-mnisthandwrittendigit-2010}, USPS \cite{291440}, and SVHN \cite{netzer2011reading} datasets. They represent a collection of images of digits from 0 to 9. %MNIST and USPS are a collection of images containing handwritten digits. SVHN is a collection of real-word images extracted from Google Street View. 
The evaluation is done by testing the model's performance on M$\rightarrow$S, U$\rightarrow$S, and M$\rightarrow$U tasks. \textbf{DomainNet.} We follow the setup used in \cite{Saito_2019_ICCV}, in which only four domains (Real (R), Clipart (C), Painting (P), and Sketch (S)) are used, with 126 classes only. 
%The defined UDA tasks focus on scenarios where the target domain is made-up of non-real images. 
The evaluated tasks are: R$\rightarrow$P, R$\rightarrow$C, R$\rightarrow$S, P$\rightarrow$C, P$\rightarrow$S, C$\rightarrow$S, and S$\rightarrow$P. In Figure \ref{fig:Benchmark}, sample images belonging to the aforementioned benchmarks are presented.
\subsection{Baseline and comparative methods}
A number of baselines have been tested in order to validate the most important components of our approach and their contributions.
\noindent
\textbf{Source Only (SO).} It is a model trained using only the labeled source images with no adaptation. 
%In our experiments, two different SO models are adopted depending on the benchmark tested. 
%For Digits, the SO model is similar to the classifier used in \cite{volpi2018adversarial} (see next subsection for details). For DomainNet benchmarks, a ResNet-101 \cite{he2016deep} model pretrained on ImageNet \cite{5206848} with the last fully connected layer as classifier, is used. 
\textbf{Adversarial Style Mining (ASM).} It is a model that also focuses  on OS-UDA. Similarly to us, ASM applies style transfer to generate target data for training. But their process requires pre-training a style generator module in an external dataset \cite{luo2020adversarial}. This is the only work with which we can directly compare to since it assumes the same starting hypotheses of our work.
\subsection{Setup}
To train our model, we used two classifier models. 
For Digits, the classifier consists of 2 blocks of a convolutional layer and a max pooling operation layer followed by three fully connected layers \cite{volpi2018adversarial}. For DomainNet, the classifier network is a ResNet-101 pretrained on the source dataset. 
The augmentation module consists of three multi-layer perceptron (MLP) networks. For the style alignment module, a VGG-16 \cite{simonyan2014very} pre-trained on ImageNet was used. In VGG-16, only the features maps from the layers $relu1\_2$, $relu2\_2$, $relu3\_3$, and $relu4\_3$ were used as we only apply the loss for the style in our experiments. %The configuration of the hyperparameters (batch size, learning rate, etc.) can be found in the supplementary material. 
\subsection{Results}
% \usepackage{multirow}
\begin{table*}[]
\centering
\caption{Classification accuracies  of the proposed TOS-UDA method on three DA tasks: MNIST to SVHN (M$\rightarrow$S), USPS to SVHN (U$\rightarrow$S), and MNIST to USP (M$\rightarrow$U). 
%The values represent the classification accuracies and standard deviations over 5 runs for each task.
}
\label{tab:da_results}
%\resizebox{0.8\textwidth}{!}{% %CVPR size
\resizebox{\textwidth}{!}{
\begin{tabular}{l|c|c|c|c|c}
\multicolumn{1}{c|}{\textbf{Model}}
& \multicolumn{1}{c|}{\textbf{Type}}
& \multicolumn{1}{c|}{\textbf{M} $\rightarrow$ \textbf{S}}                              
& \multicolumn{1}{c|}{\textbf{U} $\rightarrow$ \textbf{S}}                             
& \multicolumn{1}{c|}{\textbf{M} $\rightarrow$ \textbf{U}}
& \textbf{Average} \\
\hline
Source only
& -
& 20.30
& 15.30
& 65.40 
& 33.67 \\ 
TeachAugment \cite{Suzuki_2022_CVPR}
& -
& 31.10 $\pm$ 4.17
& 31.56 $\pm$ 2.15
& 50.26 $\pm$ 2.16 
& 37.64 \\
\hline
ASM \cite{luo2020adversarial}
& One-shot
& \textbf{46.30}
& \textbf{40.30}
& 68.00 
& 51.53 \\
OST \cite{benaim2018one}
& One-shot
& 42.50
& 34.00
& 74.80 
& 50.43 \\
Our Model
& One-shot
& 45.03 $\pm$ 1.86
& 36.67 $\pm$ 2.51
& \underline{79.45} $\pm$ 6.73
& 53.72 \\
\hline
Our Model (32 Targets)
& Few-shot
& \underline{45.88} $\pm$ 3.62
& \textbf{39.96} $\pm$ 3.63
& \textbf{79.97} $\pm$ 4.27
& \textbf{55.27}
\end{tabular}
}
\end{table*}

In Table \ref{tab:da_results}, the results on Digits are presented. The table shows the average classification accuracy and the related standard deviation, as we ran each experiment five times. 
%For comparing our method with other models, we report performance on the target domain for each adaptation task. 
Although our main objective is UDA with just one-shot, we also carried out experiments with more target samples (32). We randomly selected these samples at the start of the experiment. To calculate the style loss, each target sample was randomly paired with a sample from the current batch. For MNIST to SVHN (M$\rightarrow$S) and USPS to SVHN (U$\rightarrow$S), our approach obtains comparable results with respect to the state-of-the-art with differences lower than $1.0\%$. Although these results do not represent an improvement over the state of the art, they demonstrate that our method can achieve similar results as ASM without the need of external datasets and pretraining. For MNIST to USPS (M$\rightarrow$U), our approach in both versions (1 and 32 targets) performs better than the baseline $74.80\%$ with an increase of $+4.66\%$ and $+5.17\%$ for the 1-target and the 32-targets versions respectively. 
% \usepackage{multirow}
\begin{table*}[t]
\caption{Classification accuracies of the proposed TOS-UDA method on DomainNet across seven DA task focusing on four domains: Real (R), Clipart (C), Painting (P), and Sketch (S). 
%The values represent the classification accuracies and standard deviations resulting after averaging over 5 runs for each task.
}
\label{tab:DomainNet_results}
\resizebox{\textwidth}{!}{
%\begin{tabular}{l|cc|cc|cc|cc|cc|cc|cc|c}
\begin{tabular}{l|c|c|c|c|c|c|c|c|c}
\multicolumn{1}{c|}{\textbf{Model}}
& \multicolumn{1}{c|}{\textbf{Type}} 
& \multicolumn{1}{c|}{\textbf{R} $\rightarrow$ \textbf{C}} 
& \multicolumn{1}{c|}{\textbf{R} $\rightarrow$ \textbf{P}} 
& \multicolumn{1}{c|}{\textbf{R} $\rightarrow$ \textbf{S}} 
& \multicolumn{1}{c|}{\textbf{P} $\rightarrow$ \textbf{C}} 
& \multicolumn{1}{c|}{\textbf{P} $\rightarrow$ \textbf{R}} 
& \multicolumn{1}{c|}{\textbf{C} $\rightarrow$ \textbf{S}} 
& \multicolumn{1}{c|}{\textbf{S} $\rightarrow$ \textbf{P}} 
& \multicolumn{1}{c}{\textbf{Average}} 
\\
%\multicolumn{1}{c|}{}                                
%& \textbf{Std}                 
%& \textbf{Mean}                 
%& \textbf{Std}                 
%& \textbf{Mean}                 
%& \textbf{Std}                 
%& \textbf{Mean}                 
%& \textbf{Std}                 
%& \textbf{Mean}                 
%& \textbf{Std}                 
%& \textbf{Mean}                 
%& \textbf{Std}                 
%& \textbf{Mean}                 
%& \textbf{Std}                 
%& \textbf{Mean}         
%& 
%\\ 
\hline
Source only
& -
%& 0.79                         
& 56.59 $\pm$ 0.79                        
%& 0.50                         
& 56.79 $\pm$ 0.50                         
%& 0.86                         
& 46.25 $\pm$ 0.86                               
%& 0.83                         
& \textbf{55.55} $\pm$ 0.83                
%& 0.72                         
& \textbf{66.20} $\pm$ 0.72                
%& 1.01                         
& 52.07 $\pm$ 1.01                         
%& 1.59                         
& 44.81 $\pm$ 1.59                         
& 54.04 
\\
TeachAugment
& -
& 53.84 $\pm$ 0.56                         
& 56.70 $\pm$ 0.59                         
& 46.70 $\pm$ 1.34                               
& 50.40 $\pm$ 1.27               
& 58.64 $\pm$ 0.68               
& 50.52 $\pm$ 0.09                         
& 44.89 $\pm$ 0.83   
& 51.67
\\
\hline
ASM
& One-shot
%& 0.56                         
& 39.74 $\pm$ 0.56                         
%& 1.53                         
& 46.39 $\pm$ 1.53                        
%& 5.51                         
& 31.37 $\pm$ 5.51                         
%& 0.60                         
& 4.31 $\pm$ 0.60                         
%& 2.33                         
& 5.87 $\pm$ 2.33                          
%& 1.12                         
& 37.12 $\pm$ 1.12                         
%& 2.99                         
& 19.67 $\pm$ 2.99               
& 26.35 
\\
Our Model
& One-shot
%& 0.38                         
& \textbf{58.11} $\pm$ 0.38                
%& 0.20                         
& \textbf{58.57} $\pm$ 0.20                
%& 0.97                         
& \textbf{49.87} $\pm$ 0.97                
%& 0.62                         
& 54.24 $\pm$ 0.62                         
%& 0.32                         
& 62.72 $\pm$ 0.32                         
%& 0.25                         
& \textbf{52.88} $\pm$ 0.25                
%& 1.12                         
& \textbf{47.94} $\pm$ 1.12                
& \textbf{54.90}                                                             
\end{tabular}
}
\end{table*}
In Table \ref{tab:DomainNet_results}, we present results for DomainNet. We compare our model against three baselines: Source only (SO), TeachAugment and Adversarial Style Mining (ASM). We generate the results for these three baselines given that to the best of our knowledge is the first time that the DomainNet benchmark is used for OS-UDA. 
%The DomainNet benchmark has seven adaptation tasks. 
We report the accuracy and standard deviation for each task and their average across all tasks.
%We present the classification accuracy for each task individually as well as their standard deviations and the averaged accuracy over all the tasks. 
Compared to ASM, our approach obtains higher accuracy in all seven DA tasks, setting the SoTA for five of these tasks. For P$\rightarrow$C and P$\rightarrow$R, our approach performs poorly by having results similar to the SO baseline. This indicates that for some domains, the augmentation module may struggle to learn transformations for the domain complex style from just 1 target.
\subsection{Ablation analysis}
%%% New text
We conducted ablations to analyze the contribution of the augmentation and style alignment modules. Specifically, we removed the loss for style alignment and the adversarial loss component in two separate experiments. Table \ref{tab:ablation_results} shows the performance of our model compared to a pre-trained SO model and our method with excluded losses. Our method outperformed the SO model even with only one of the losses guiding the augmentations. However, the best results were achieved when both losses were used, confirming the contribution of both modules.
%\usepackage{multirow}
\begin{table}[]
\centering
\caption{Ablation analysis of the proposed TOS-UDA approach  on Digits benchmark}
\label{tab:ablation_results}
%\resizebox{\columnwidth}{!}{% %CVPR size
%\resizebox{0.75\textwidth}{!}{% %CVPR size
\begin{tabular}{l|c|c|c}
\multicolumn{1}{c|}{\textbf{Model Setup}} 
& \multicolumn{1}{c|}{\textbf{M} $\rightarrow$ \textbf{S}}                              
& \multicolumn{1}{c|}{\textbf{U} $\rightarrow$ \textbf{S}}                             
& \multicolumn{1}{c}{\textbf{M} $\rightarrow$ \textbf{U}} \\
\hline
Source only
& \multicolumn{1}{c|}{20.30}
& \multicolumn{1}{c|}{15.30}
& \multicolumn{1}{c}{65.40} \\
\hline
Our w/o style. loss        
& \multicolumn{1}{c|}{ 22.43 $\pm$ 1.71}
& \multicolumn{1}{c|}{ 28.45 $\pm$ 1.02}
& \multicolumn{1}{c}{ 74.74 $\pm$ 3.01} \\
Our w/o classif. loss       
& \multicolumn{1}{c|}{20.55 $\pm$ 1.28}
& \multicolumn{1}{c|}{20.93 $\pm$ 2.93}
& \multicolumn{1}{c}{71.22 $\pm$ 10.48} \\
\hline
Our method        
& \multicolumn{1}{c|}{45.03 $\pm$ 1.86}
& \multicolumn{1}{c|}{36.67 $\pm$ 2.51}
& \multicolumn{1}{c}{79.45 $\pm$ 6.73} 
\end{tabular}
%}
\end{table}

%%% Previous text
%We performed ablations to analyze the contribution of augmentation and style alignment modules.  We performed experiments, 1) by removing the loss for style alignment, and 2) by removing the adversarial loss component. In Table \ref{tab:ablation_results}, we report the performance of our model with 1) a SO model which is pre-trained only on source images, and 2) Models trained with our method where we excluded the style loss and adversarial classification loss. We observed that our method results in better performance than the SO model even when only style loss, or only adversarial loss is used to guide the augmentations. As expected, our method performed better when both losses were used, thus confirming the contribution of both style alignment module and adversarial augmentation modules.
%
\section{Conclusions}
In this paper, we present our novel Target-driven One-Shot Unsupervised Domain Adaptation (TOS-UDA) approach. Our method focuses on solving OS-UDA, which is the problem of performing domain adaptation with only one unlabeled target sample. Our method gradually augments the source samples to match them in style with the available target. To guide the augmentation process, adversarial training was used to encourage the augmentation module to learn diverse augmentations, while reaching a style similarity to the target sample. In this way, our augmentation module can learn effective transformations for OS-UDA. Further, our proposed training pipeline is simpler and end-to-end as compared to existing OS-UDA method such as ASM \cite{luo2020adversarial}. 
%Moreover, our method does not depend on additional training efforts such as pretrained style transfer modules or additional datasets.
We tested our approach for image classification tasks in two well-known domain adaptation benchmarks: Digits and DomainNet. We demonstrated that our method performs better that the selected baselines in almost all the DA tasks. 
In the future, we will focus on extending the augmentation module to perform more specialized augmentations as the basic geometric and color augmentations may not be sufficient to simulate the style complexity of certain domains. %\VM{UNCLEAR}
\bibliographystyle{splncs04}
\bibliography{egbib}
\end{document}